%% file: main.tex
\documentclass[12pt]{msml2021} 

\input{abbrv.tex}
\input{preamble.tex}

\makeatletter
\let\Ginclude@graphics\@org@Ginclude@graphics
\makeatother

\title[Hybrid Regularization for Random Feature Models]{Avoiding The Double Descent Phenomenon of Random Feature Models Using Hybrid Regularization}
\usepackage{times}

\msmlauthor{\Name{Kelvin Kan} \Email{kelvin.kan@emory.edu}\\
 \Name{James G Nagy} \Email{jnagy@emory.edu}\\
 \Name{Lars Ruthotto} \Email{lruthotto@emory.edu}\\
 \addr Department of Mathematics, Emory University, 400 Dowman Dr, 30322 Atlanta, GA, USA}

\begin{document}

\maketitle

\begin{abstract}
We demonstrate the ability of hybrid regularization methods to automatically avoid the double descent phenomenon arising in the training of random feature models (RFM). The hallmark feature of the double descent phenomenon is a spike in the regularization gap at the interpolation threshold, i.e. when the number of features in the RFM equals the number of training samples. To close this gap, the hybrid method considered in our paper combines the respective strengths of the two most common forms of regularization: early stopping and weight decay. The scheme does not require hyperparameter tuning as it automatically  selects the stopping iteration and weight decay hyperparameter by using generalized cross-validation (GCV). This also avoids the necessity of a dedicated validation set. While the benefits of hybrid methods have been well-documented for ill-posed inverse problems, our work presents the first use case in machine learning. To expose the need for regularization and motivate hybrid methods, we perform detailed numerical experiments inspired by image classification. In those examples, the hybrid scheme successfully avoids the double descent phenomenon and yields RFMs whose generalization is comparable with classical regularization approaches whose hyperparameters are tuned optimally using the test data. We provide our MATLAB codes for implementing the numerical experiments in this paper at \url{https://github.com/EmoryMLIP/HybridRFM}.
\end{abstract}

\begin{keywords}%
  Double descent, random feature model, generalization error, hyperparameter selection, regularization
\end{keywords}

\section{Introduction}

Random feature models (RFM)~\citep{RahimiRecht2007} are machine learning models that express the relationship between given input and output features as the concatenation of a random feature extractor and a linear model whose weights are optimized. 
The double descent phenomenon relates the generalization error of the RFM to the number of random features. 
It was observed experimentally for least-squares fitting problems in~\cite{belkin2019,advani2020}.
In short, it manifests as a decreasing generalization error as the number of random features grows, a sharp spike when it reaches the number of training samples, followed by a decaying generalization error as the number of features is increased further.

The theoretical understanding of the double descent phenomenon has matured considerably in the last few years.  In particular, \cite{belkin2019} studied the phenomenon with a data-fitting perspective. In \cite{advani2020,MaEtAl2020}, the generalization dynamics were analyzed for gradient flows and early stopping was identified as a practical remedy; however, the success of this regularization requires a judicious choice of the stopping time by the user. Our scheme overcomes this obstacle by automatically selecting all regularization hyperparamters. \cite{belkin2019b,hastie2019surprises} provided a statistical analysis of the double descent phenomenon. \cite{mei2019} performed a rigorous analysis of the asymptotic behavior of the test errors.
The phenomenon also arises in more general settings, for example, for logistic regression loss functions~\citep{deng2019model,kini2020analytic} and neural network models trained with hinge losses~\citep{geiger2020}.

In this paper, we provide an inverse problems perspective and show that the double descent phenomenon can be explained by the ill-posedness of the learning problem.
Hence, one can avoid it by early stopping and weight decay regularization - provided an effective choice of hyperparameters; we therefore extend similar arguments made , e.g., by \cite{advani2020,MaEtAl2020}. 

As a new remedy to tackle the double descent phenomenon, we present a computationally efficient hybrid method~\citep{chung2008} that automatically tunes its hyperparameters and avoids the double descent phenomenon. 
The development of hybrid methods has been a fruitful and important direction in inverse problems recently~\citep{chung2015,gazzola2015,chung2017}.
We use the hybrid method implemented in~\cite{Gazzola:2018cc}, which combines the respective strengths of early stopping and weight decay. 
Specifically, the method performs a few iterations of the numerically stable Krylov subspace method LSQR~\citep{paige1982,paige1982b} and adaptively selects the weight decay parameter at each iteration using generalized cross-validation (GCV)~\citep{golub1979}. 
Notably, the scheme does not require a dedicated validation set.
The effectiveness and the scalability of the method have been documented for various large-scale imaging problems~\citep{chung2008,chung2015,chung2017}. 

To our knowledge, our paper is the first to demonstrate the benefits of hybrid methods for training random feature models. 
We perform extensive experiments using the MNIST and CIFAR datasets and compare the effectiveness of early stopping, weight decay, and hybrid regularization for small, medium, and large RFM models.
In those examples, the hybrid scheme successfully avoids the double descent phenomenon. Also, it yields RFMs whose generalization error is comparable with that of classical regularization approaches even when their hyperparameters are tuned to minimize the test loss, a procedure that is to be avoided in realistic application. To enable reproducibility, we provide our codes used to perform the experiments in this paper at \url{https://github.com/EmoryMLIP/HybridRFM}.

The remainder of the paper is organized as follows.
In \Cref{sec:DD}, we review the training problem in RFMs, describe the double descent phenomenon, and relate it to the ill-posedness of the training problem. 
In \Cref{sec:reg}, we demonstrate that - with a proper choice of hyperparameters - early stopping and weight decay can effectively regularize the RFM training and avoid the double descent phenomenon. 
In \Cref{sec:hybrid}, we present the hybrid scheme and show its ability to avoid the double descent phenomenon without requiring hyperparameter tuning.
In \Cref{sec:discussion}, we discuss and compare the numerical results of the different schemes before concluding the paper in \Cref{sec:conclusion}.

\section{The Double Descent Phenomenon}\label{sec:DD}
We describe the double descent phenomenon, which occurs in the training of the random feature model (RFM)~\citep{RahimiRecht2007}. 
We discuss the problem setup of RFM, introduce the data used in our experiments, explain the double descent phenomenon, and relate the double descent phenomenon to ill-posed inverse problems.

\paragraph{Problem setup}
We consider a supervised learning problem where we are given the matrix of input features $\bfY \in \mathbb{R}^{n\times n_{\rm f}}$  and the matrix of corresponding outputs $\bfC \in \mathbb{R}^{n\times n_{\rm c}}$.
Here, $n_{\rm f}$ is the number of input features and $n_{\rm c}$ is the number of output features (e.g., the number of classes) and the $n$ examples are stored row-wise. 
The idea in RFM is to transform the input features by applying a random nonlinear transformation $f : \mathbb{R}^{n_{\rm f}}\to\mathbb{R}^m$ to each row in $\bfY$ and then train a linear model to approximate the relationship between $f(\bfY)$ and $\bfC$.
Here, the transformation $f(\bfY)$ is applied row-wise and yields a new representation of the features in $\mathbb{R}^{n \times m}$.
The dimension $m$ controls the expressiveness of the RFM and can be chosen arbitrarily; generally, larger values of $m$ increase the expressiveness of the RFM.

Similar to~\cite{MaEtAl2020} we define our RFM using a randomly generated matrix $\bfK \in \mathbb{R}^{n_{\rm f}\times (m-1)}$, a bias vector $\bfb \in \mathbb{R}^{m-1}$, and an activation function $a : \mathbb{R}\to\mathbb{R}$, as
\begin{equation}
    \bfZ = f(\bfY) =  \left[
        \begin{array}{ll}
            a(\bfY \bfK + {\bf 1}_n \bfb^\top) & {\bf 1}_n
        \end{array}
    \right],
\end{equation}
where the activation function is applied element-wise, and ${\bf 1}_n \in \mathbb{R}^n$ is a vector of all ones used to model a bias term.

The RFM training consists of finding the linear transformation $\bfW \in \mathbb{R}^{m \times n_{\rm c}}$
such that $\bfZ \bfW \approx \bfC$. 
As in~\cite{MaEtAl2020}, we measure the quality of the model using the least squares loss function.
The double descent phenomenon arises when the weights are obtained by solving the unregularized regression problem, i.e., 
\begin{equation}\label{eq:LS_noreg}
   \bfW^*_{\rm LS} \in  \argmin_{\bfW \in \mathbb{R}^{m\times n_{\rm c}}} \frac{1}{2n} \| \bfZ \bfW - \bfC\|_{\rm F}^2,
\end{equation}
where $\| \cdot \|_{\rm F}$ is the Frobenius norm.
Problem~\eqref{eq:LS_noreg} is separable, which means that it can be decoupled into $n_{\rm c}$ least-squares problems each of which determines one column of $\bfW^*_{\rm LS}$.
Therefore, without loss of generality, we focus our discussion on the case $n_{\rm c}=1$ and consider the problem 
\begin{equation}\label{eq:LS_noreg_sep}
    \bfw_{\rm LS}^* \in \argmin_{\bfw \in \mathbb{R}^m}  \frac{1}{2n} \| \bfZ \bfw  - \bfc\|_{2}^2.
\end{equation}
Here, $\bfc \in \mathbb{R}^n$ are the output labels. 
For example, when choosing $\bfc$ as the $i$th column $\bfC$ the solution of~\eqref{eq:LS_noreg_sep} is  the $i$th column of $\bfW^*_{\rm LS}$. 

The actual goal in learning is not necessarily to optimally solve~\eqref{eq:LS_noreg_sep} but to  obtain a RFM model that generalizes beyond the training data. 
To gauge the generalization of the model defined by $\bfw^*_{\rm LS}$, consider the test data set given by $\bfY_{\rm test} \in \mathbb{R}^{n_{\rm test} \times n_{\rm f}}$ and $\bfC_{\rm test} \in \mathbb{R}^{n_{\rm test} \times n_{\rm c}}$. 
Then, the model defined by $\bfw^*_{\rm LS}$, $\bfK$, and $\bfb$ generalizes well if the generalization gap
\begin{equation}
    \frac{1}{2 n_{\rm test}} \| f(\bfY_{\rm test}) \bfw_{\rm LS}^*  - \bfc_{\rm test}\|_{2}^2 -  \frac{1}{2 n} \| f(\bfY) \bfw_{\rm LS}^*  - \bfc\|_{2}^2
\end{equation}
is sufficiently small. 
The main goal of our paper is to understand why solutions to~\eqref{eq:LS_noreg_sep} may fail to generalize and how to regularize~\eqref{eq:LS_noreg_sep} such that its solution reliably approximates the input-output relation for unseen data.

\begin{figure}[t]
\floatconts
  {fig:double_descent}
  {\caption{(a)-(b): The double descent phenomenon observed in random feature model on MNIST and CIFAR10 data. (c)-(e): The Picard plot for $\bfZ$ and $\bfc$ on CIFAR10 data with $n=1024$ and $m=512$ (overdetermined), $1024$ (unique) and $2048$ (underdetermined). All the values are averaged over 5 random trials. The top, middle and bottom  curves are $\sigma_j$, $| \bfu_j^\top \bfc |$ and $| \bfu_j^\top \bfc|/\sigma_j $ defined in~\eqref{eq:SVD_psuedoinverse}, respectively. When $m=n$, there is a plummet in $\sigma_j$ at the end, and it renders $| \bfu_j^\top \bfc|/\sigma_j $ large. These large values dominate the optimal solution $\bfW$. Thus, there is a spike in the norm of $\bfW$ and hence the testing loss.}}
  {\includegraphics[width=1\textwidth]{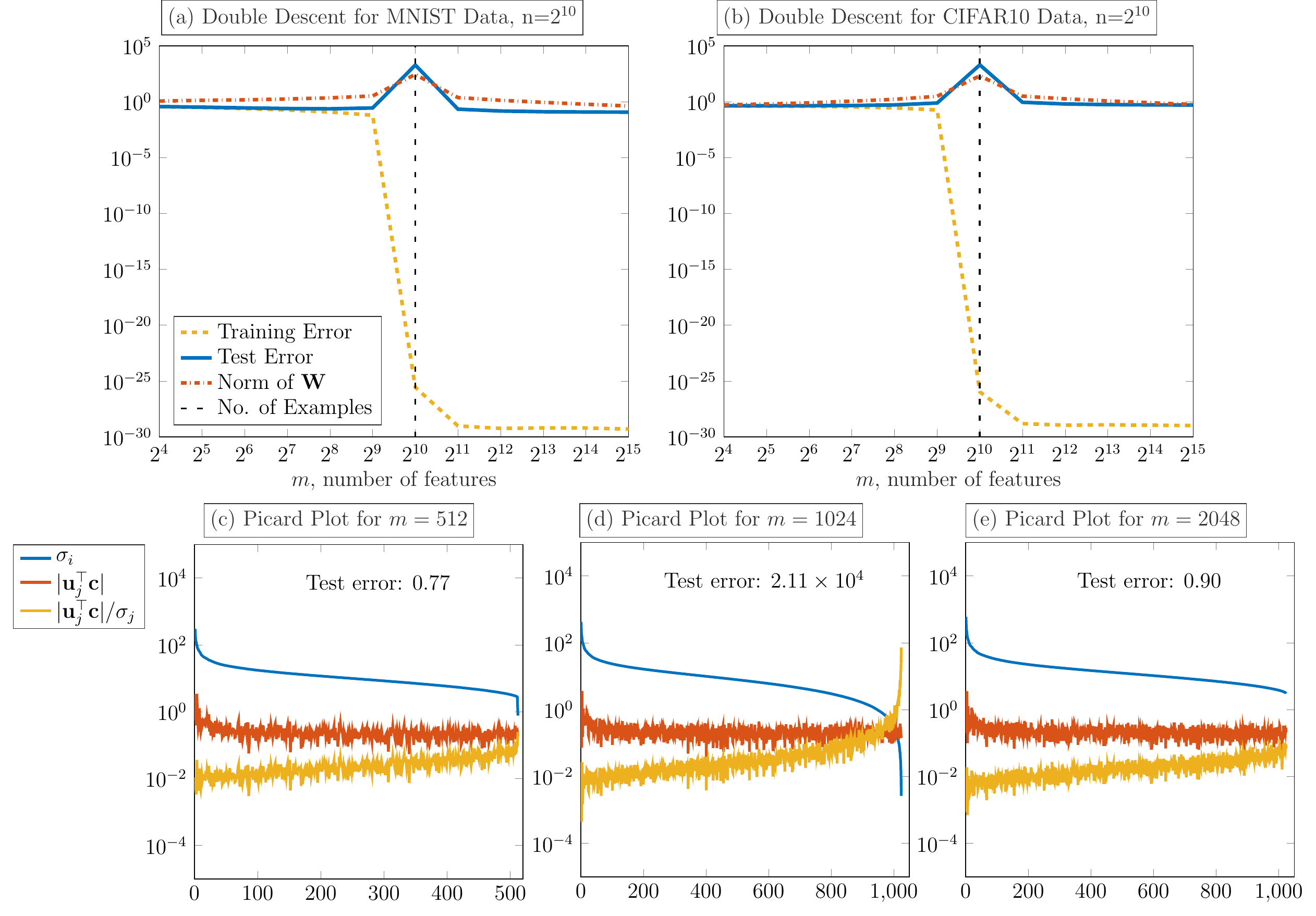}}
  \end{figure}
  
  \paragraph{Examples from image classification}
Throughout the paper, we will use two common image classification benchmarks to illustrate the techniques. 
Specifically, we use the MNIST dataset consisting of $28\times 28$ gray scale images of hand-written digits~\citep{LeCun1990} and the CIFAR 10 dataset~\citep{CIFAR} consisting of $32\times32$ RGB images of objects divided into one of ten categories.
From each dataset we randomly sample $n=1,024$ training images and their labels. 

Each dataset also contains $n_{\rm test} = 10,000$ labeled test images, which we use to compute the generalization gap of the trained model.
To obtain competitive baseline results for our method we also use the test images to optimize the hyperparameters of state-of-the-art methods.
It is important to emphasize that our hybrid method selects hyperparametes for regularization and stopping criteria automatically and hence does not use the test data.

In our experiments, we use the ReLU activation function $a(x) = \max(x,0)$ and generate $\bfK$ and $\bfb$ using a uniformly random distribution drawn from a unit sphere; see also~\cite{MaEtAl2020}.
  
\paragraph{The Double Descent Phenomenon}
The key hyperparameter of an RFM is the dimensionality of the feature space, $m$.
The double descent phenomenon~\citep{belkin2019,MaEtAl2020,geiger2020} is observed in the generalization gap for different choices $m$. 
This behavior can be described in three stages: when $m<n$, $m=n$ and $m>n$, see~\Cref{fig:double_descent}(a)-(b). In the following, we explain these three stages using a data fitting viewpoint. 
\begin{itemize}
    
\item When $m<n$, the learning problem~\eqref{eq:LS_noreg_sep} is overdetermined. That is, there are more equations than variables in $\bfZ \bfw = \bfc$. There is no solution to perfectly describe the input-output relation in general. Hence as $m$ increases, we are able to fit both the training and test data better and the generalization gap decreases.

\item When $m=n$, $\bfZ$ is square and, in our experience, invertible. 
Thus the optimal solution to~\eqref{eq:LS_noreg_sep} is unique and satisfies $\bfZ \bfw = \bfc$. 
In order words, the training data can be fitted perfectly, and the training loss is essentially zero. 
However, the uniqueness of the optimal $\bfw$ also implies that we have no choice but to perfectly fit to the weakly present features in $\bfZ$, which are not relevant to the classification. The perfect loss on the training data combined with an increase in test loss then causes a spike in the generalization gap. 

\item When we further increase $m$ such that $m>n$, the problem is underdetermined, and there are infinitely many optimal $\bfw$ to achieve an objective function value of zero. 
It has been observed that selecting the solution with the minimal norm reduces the risk of fitting weakly present features; see, e.g., \cite{belkin2019}. Therefore, generally, the generalization gap decreases as $m$ grows.
\end{itemize}

\paragraph{Double Descent and Ill-Posedness}
To better understand and avoid the double descent phenomenon, we use its connections to ill-posed inverse problems~\citep{engl1996,hansen1998,hansen2010}. 
Consider the singular value decomposition (SVD) $\bfZ=\bfU {\bf \Sigma }\bfV^\top$.
Here, $\bfU=[\bfu_1,\bfu_2,...,\bfu_n]$ and $\bfV=[\bfv_1,\bfv_2,...,\bfv_m]$ are orthogonal matrices, and $\boldsymbol{\Sigma} \in \mathbb{R}^{n \times m}$ contains the singular values $\{ \sigma_j \}_{j=1}^{\text{min}(m,n)}$ in descending order on its diagonal and is zero otherwise. 
Let $r$ be the rank of $\bfZ$, that is, the last index such that $\sigma_r>0$. 
When the singular values decay to zero smoothly and without a significant gap, it is common to call problem ~\eqref{eq:LS_noreg_sep} ill-posed and a large generalization error has to be expected for some labels $\bfc$.

Using the SVD, the minimum norm solution of~\eqref{eq:LS_noreg_sep}, can be written explicitly as
\begin{equation}\label{eq:SVD_psuedoinverse}
    \bfw^*_{\rm LS} =\sum_{j=1}^{r} \frac{\bfu_j^\top \bfc}{\sigma_j} \bfv_j.
\end{equation}
This formulation shows that the contribution of $\bfv_j$ to $\bfw^*_{\rm LS}$ is scaled by the ratio between $\bfu_j^\top \bfc$ and $\sigma_j$. 
If this ratio is large in magnitude then the solution is highly sensitive to perturbations of the data $\bfc$ along the direction $\bfu_j$ and the corresponding singular vector $\bfv_j$ gets amplified in the solution.
Also, $\sigma_j$ quantifies the importance of the feature $\bfu_j$ in the data matrix $\bfZ$.
Therefore, intuitively one wishes $|\bfu_j^\top \bfc|$ to decay as $j$ grows. 
In other words, this observation also suggests that for ill-posed problems the generalization gap depends on the decay of $|\bfu_j^\top \bfc|$.

We illustrate the quantities in~\eqref{eq:SVD_psuedoinverse} for the three stages of the double descent in Figures~\ref{fig:double_descent}(c)-(e) using the CIFAR10 example.
Here, we plot $\sigma_j$, $|\bfu_j^\top \bfc|$ and $|\bfu_j^\top \bfc|/\sigma_j$ in Figures~\ref{fig:double_descent}(c)-(e); the resulting plot is known as a Picard plot~\citep{hansen2010}.
From the decay of the singular values (see the blue line), we see that $m=n$ leads to an ill-posed problem as the  $\sigma_j$ decay to zero with no significant gap. 
Also, for $m=n$, the magnitude of $\bfu_j^\top \bfc$ (see red line) remains approximately constant.
This combination causes a surge of $|\bfu_j^\top \bfc/\sigma_j|$ (see yellow line), which causes an increase of the norm of $\bfw$; see the red dashed line in~\Cref{fig:double_descent}(a)-(b). 
When $m\neq n$ the problem is not ill-posed as the decay of the singular values is less pronounced. 
This correlation between the ill-posedness and the observed double descent phenomenon, motivates us to employ state-of-the-art techniques for regularizing ill-posed inverse problems to improve the generalization of random feature models. 

\section{Regularization By Early Stopping and Weight Decay}\label{sec:reg}
Regularization techniques are commonly used to improve the generalization of machine learning models (see, e.g.,  \citep[Chapter 7]{goodfellow2016}) and to enhance the solution of ill-posed inverse problems (see, e.g., \cite{engl1996,hansen1998, hansen2010}).
Despite differences in notation and naming, the basic ideas in both domains are similar. 
This section aims to provide the background of the two most common forms of regularization: early stopping  and weight decay, which, in the inverse problems literature, are better known as iterative and Tikhonov regularization, respectively.
While the techniques in this section are standard, using hybrid approaches to train random feature models, as we discuss in the next section, is a novelty of our work.

Using the SVD of the feature matrix $\bfZ = \bfU \boldsymbol{\Sigma}{\bfV}^\top$, we can write and analyze most regularization schemes for~\eqref{eq:LS_noreg_sep} using their corresponding filter factors $\phi_j$ that control the influence of the $j$th term in~\eqref{eq:SVD_psuedoinverse}.
To be precise, the regularized solutions can be written as
\begin{equation}\label{eq:sol_RLS}
    \bfw_{\rm reg}=\sum_{j=1}^{r} \phi_{j} \frac{\bfu_j^\top \bfc }{\sigma_j} \bfv_j.
\end{equation}
A simple example is the truncated SVD, in which terms associated with small singular values are ignored by using the filter factors
\begin{equation}
\phi_{{\rm TSVD},j}(\tau) = 
    \begin{cases}
        1, & \sigma_j > \tau \\
        0, & \sigma_j \leq \tau
    \end{cases},
\end{equation}
where the choice of $\tau \geq 0$ is crucial to trade off the reduction of the training loss and the regularity of the solution, which is needed to generalize. In the remainder of the section, we briefly review early stopping and weight decay. 
\paragraph{Early Stopping}
A common observation during the training of random feature models with iterative methods is an initially sharp decay of both the training and test losses followed by a widening of the generalization gap in later iterations. 
For least-squares problems such as~\eqref{eq:LS_noreg_sep} this behavior, also known as semiconvergence, typically arises when the iterative method converges quicker on the subspace spanned by the singular vectors associated with large singular values than on those associated with small singular values.
A straightforward and popular way to regularize the problem then is to stop the iteration early; see, e.g., \cite{MaEtAl2020} for RFMs,~\cite{yao2007} for neural networks, and~\cite{chung2008} for image recovery.

As a simple example to show the regularizing effect of early stopping, we consider the gradient flow (GF) applied to~\eqref{eq:LS_noreg_sep}, which reads
\begin{equation}
    \partial_t \bfw_{\rm GF}(t) = - \frac{1}{n} \bfZ^\top(\bfZ \bfw_{\rm GF}(t)  - \bfc) , \quad \bfw_{\rm GF}(0) = {\bf 0}.
\end{equation}
As also shown in \cite{MaEtAl2020}, when the SVD of the feature matrix is available, $\bfw_{\rm GF}(t)$ can be computed via ~\eqref{eq:sol_RLS} using the filter factors
\begin{equation}\label{eq:phiGF}
    \phi_{{\rm GF},j} (t) = 1-e^{-\sigma_j^2 t/(mn)}.
\end{equation}
From this observation, we can see that as $t$ grows, the filter factors converge to one, and $\bfw_{\rm GF}(t)$ converges to the solution of the unregularized problem~\eqref{eq:LS_noreg_sep}. 
Furthermore, we see that for any fixed time, the filter factors decay as $j$ grows, which reduces the sensitivity to perturbations of the data $\bfc$ along the directions associated with small singular values.
  \begin{figure}[t]
\floatconts
  {fig:GF_WD}
  {\caption{(a)-(b): The results of applying gradient flow (GF) to~\eqref{eq:LS_noreg} when $m=n=1024$ at different time $t$. (c)-(d): The results obtained from weight decay~\eqref{eq:RLS_sep} with different  hyperparameter $\alpha$. The optimal stopping time/regularization hyperparameter is highlighted. We can see that both methods exhibit a semiconvergence behavior. In particular, their generalizability depends on the choice of hyperparameters, which varies form problem to problem and has to be make judiciously. Determining the optimal hyperparameters requires access to the test data. Yet, in practice test data are not allowed to use for training.}}
  {\includegraphics[width=.8\textwidth]{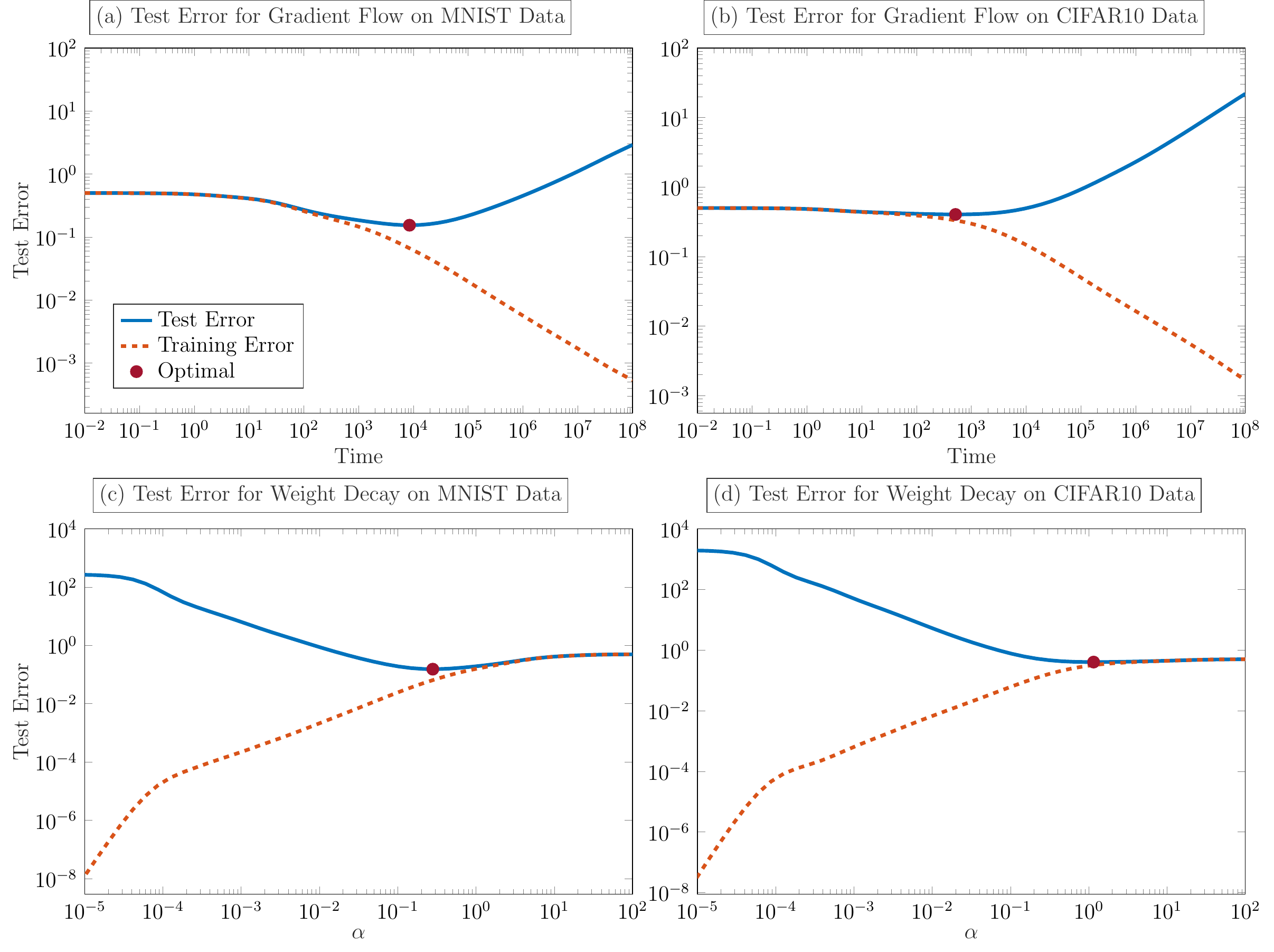}}
  \end{figure}
In the top row of~\Cref{fig:GF_WD}, we show numerical results for the early stopping applied to our two test problems.
The qualitative behavior is comparable for both datasets: Initially, both training and test losses decay with no noticeable gap but at later times, the test losses increases dramatically.
A difference between the two datasets is that the optimal stopping time (i.e., the time with the smallest test loss) differs by about two orders of magnitude.
Hence, the stopping time is the key hyperparameter that needs to be chosen judiciously and depends on the problem. 
Determining an effective stopping time is even more difficult in realistic applications, when this decision must not be based on the test data set.

In addition to gradient descent and its stochastic versions, many other first-order optimization methods enjoy similar regularizing properties. 
In particular, Krylov subspace methods have been commonly used in ill-posed inverse problems due to their  superior convergence properties; see, e.g., ~\cite{hansen1998,hansen2010}.

The cross-validation of early stopping is easy to perform, for example, one can use a part of the training data for validation and stop the training process when the validation error is minimized. However, such an approach reduces the number of training data. The regularization properties and their analysis also depend heavily on the underlying iterative method. Moreover, early stopping generally prefers slowly converging schemes to have a broader range of optimal stopping points.

\paragraph{Weight Decay}
The idea in the direct regularization methods, known as weight decay in machine learning~\citep[Chapter 7]{goodfellow2016} and Tikhonov regularization~\cite{engl1996,hansen1998} in inverse problems, 
is to add an extra term to the objective in~\eqref{eq:LS_noreg_sep} such that the solution to the obtained regularized problem generalizes well. 
There are many options for choosing regularization terms.
In our work, we consider the squared Euclidean norm of $\bfw$ and define the regularized solution as
\begin{equation}\label{eq:RLS_sep}
    \bfw_{\rm WD}(\alpha) = \argmin_{\bfw} \frac{1}{2n} \| \bfZ \bfw  - \bfc\|_{2}^2 + \frac{\alpha^2}{2}  \| \bfw\|_{2}^2.
\end{equation}
Here, the hyperparameter $\alpha \geq 0$ trades off minimization of the loss and the norm of $\bfw_{\rm WD}$.
An advantage compared to early stopping is that any iterative or direct method used to solve~\eqref{eq:RLS_sep} will ultimately provide the same solution.

Using the SVD of $\bfZ$ we can see that $\bfw_{\rm WD}(\alpha)$ can be computed using~\eqref{eq:sol_RLS} and the filter factors
\begin{equation}\label{eq:phiWD}
    \phi_{{\rm WD},j}(\alpha)=\frac{\sigma_j^2}{\sigma_j^2 + n\alpha^2},
\end{equation}
which are also called the Tikhonov filter factors~\citep{hansen1998}. 
We can see that when $\alpha$ is chosen relatively small, the filter factors associated with large singular values remain almost unaffected while those corresponding to small singular values may be close to zero.  Thus, similar to TSVD and early stopping, weight decay can reduce the sensitivity of $\bfw_{\rm WD}$ to perturbations of the data along the directions associated with small singular values.  

We invesigate the impact of choosing $\alpha$ on the generalization in a numerical experiment for the  MNIST and CIFAR10 example; see~\Cref{fig:GF_WD}(c)-(d).
The qualitative behavior is comparable for both datasets: as $\alpha$ increases the training error increases monotonically, while the test error first decays and then finally grows.
Due to this semiconvergence, a careful choice of $\alpha$ can improve generalization; we visualize the  hyperparameter $\alpha$ that yields the lowest test loss with red dots.
A key difference between the examples is that the optimal values of $\alpha$ differs by about one order of magnitude, which highlights its problem-dependence. As in early stopping, we re-iterate that the test data must not be used to select the optimal value of $\alpha$. 

The solution of weight decay has a simple representation~\eqref{eq:phiWD}. This renders its analysis simple. Moreover, the same solution is obtained regardless of the choice of solvers. Thus, in contrast to early stopping, the most efficient scheme can be used. Yet the optimal choice of the hyperparameter $\alpha$ requires cross-validation~\citep{kohavi1995} in which the problem has to be solved many times.
\section{Hybrid Regularization: The Best of Both Worlds}\label{sec:hybrid}

Hybrid methods belong to the most effective solvers for ill-posed inverse problems and have been widely used, for example, in large-scale image recovery~\citep{chung2008,chung2015,chung2017}. 
The key idea in hybrid methods is to combine the respective advantages of early stopping and weight decay while avoiding their disadvantages.
In this section, we briefly review the technique \texttt{IRhybrid\_lsqr} from the open source MATLAB package~\cite{IRtools} and for the first time apply the scheme in the machine learning domain for training random feature models.
This hybrid method employs LSQR~\citep{paige1982,paige1982b} that at each iteration projects the regularized least-squares problem~\eqref{eq:RLS_sep} onto a small-dimensional subspace and adaptively selects the weight decay parameter using generalized cross-validation (GCV)~\citep{golub1979}.
The resulting hybrid method does not involve any hyperparameter tuning and, in our experiments, successfully avoids the semiconvergence, hence, the double descent phenomonon.

\paragraph{LSQR Algorithm} 
 LSQR~\citep{paige1982,paige1982b} is an iterative method for solving (regularized) least-squares problems.
With comparable computational costs per iteration, the numerical stability and convergence of LSQR generally is superior to gradient descent, particularly for ill-posed problems. 
The $k$th iteration of LSQR solves the projection of~\eqref{eq:RLS_sep} onto the $k$-dimensional Krylov subspace $\mathcal{K}_k = {\rm span} \{\bfZ^\top\bfc,(\bfZ^\top\bfZ)\bfZ^\top\bfc, \ldots, (\bfZ^\top\bfZ)^{k-1}\bfZ^\top\bfc\}$.
This projection is obtained using  Lanczos bidiagonalization~\citep{golub1965} of the feature matrix $\bfZ$ with the initial vector $\bfc$, which reads
\begin{align}
    \bfZ^\top \bfQ_k &= \bfP_k \bfB_k^\top + \gamma_{k+1} \bfp_{k+1} \bfe^\top_{k+1}, \label{eq:LB1}\\
    \bfZ \bfP_k &= \bfQ_k \bfB_k, \label{eq:LB2}
\end{align}
where $\bfQ_k\in \mathbb{R}^{n \times (k+1)}$ and $\bfP_k\in \mathbb{R}^{m \times k}$ have orthonormal columns, $\bfB_k\in \mathbb{R}^{(k+1)\times k}$ is a lower bidiagonal matrix, $\bfe_{k+1}\in\mathbb{R}^{k+1}$ is the $(k+1)$th standard basis vector, and  $\gamma_{k+1}$ and $\bfp_{k+1}$ will be the $(k+1)$th diagonal entry of $\bfB_{k+1}$ and the $(k+1)$th column of $\bfP_{k+1}$, respectively.


Using the bidiagonalization, we derive the projection of~\eqref{eq:RLS_sep} as follows
\begin{align}
    \min_{\bfx \in \mathcal{K}_k} \frac{1}{2n} \| \bfZ \bfx  - \bfc\|_{2}^2 + \frac{\alpha^2}{2}\|\bfx\|^2 &= \min_{\bff \in\mathbb{R}^k} \frac{1}{2n} \| \bfZ \bfP_k \bff  - \bfc\|_{2}^2 + \frac{\alpha^2}{2} \|\bff\|^2 \\
    \intertext{where we used that the columns of $\bfP_k$ form an orthonormal basis of $\mathcal{K}_k$, which also implies that $\|\bfP_k\bff\|=\|\bff\|$. Next, using~\eqref{eq:LB2} gives}
    &= \min_{\bff\in\mathbb{R}^k} \frac{1}{2n} \| \bfQ_k \bfB_k \bff  - \bfc\|_{2}^2 + \frac{\alpha^2}{2} \|\bff\|^2 \\
    \intertext{Using the orthonormality of the columns of $\bfQ_k$ and the fact that $\bfQ_k$ contains $\frac{\bfc}{\| \bfc \|_2}$ in its first column, we obtain the projected problem}
    &= \min_{\bff\in\mathbb{R}^k} \frac{1}{2n} \|  \bfB_k \bff  - \beta \bfe_1\|_{2}^2 + \frac{\alpha^2}{2} \|\bff\|^2, \label{eq:proj_eq_LS}
\end{align}
where $\beta=\| \bfc \|_2$ and $\bfe_1 \in \mathbb{R}^{k+1}$ is the first standard basis vector. Here, the $k$-dimensional projected problem~\eqref{eq:proj_eq_LS} is greatly reduced in size compared to the original $m$-dimensional problem~\eqref{eq:RLS_sep}. The $k$th iteration of LSQR is the solution to the projected problem and can be computed using the regularized pseudoinverse $\bfB_{k,\alpha}^\dagger$ via
\begin{equation}\label{eq:projected_sol}
    \bfP_k \bff_\alpha = \beta \bfP_k\bfB_{k,\alpha}^\dagger \bfe_1 
    \quad \text{ with } \quad \bfB_{k,\alpha}^\dagger = \frac{1}{n} \left(\frac{1}{n}\bfB_k^\top \bfB_k + \alpha^2 \bfI\right)^{-1} \bfB_k^\top.
\end{equation}

\paragraph{Automatic Weight Decay} 
In weight decay, the choice of the hyperparameter $\alpha$ is crucial in order to successfully filter the small singular values. One straight forward way to choose it is to perform cross-validation. Having the small-dimensional projected problem~\eqref{eq:proj_eq_LS} allows us to test multiple candidate $\alpha$'s for cross-validation efficiently. Here, the parameter selection can be done even more effectively by using statistical criteria such as generalized cross-validation (GCV)~\citep{golub1979,engl1996,hansen1998,vogel2002}, weighted GCV~\citep{chung2008}, L-Curve~\citep{calvetti1999} and discrepancy principle~\citep{vogel2002}. In this paper, we use GCV for simplicity. 

The idea of GCV is to pick a weight decay hyperparameter that gives good generalization power. Specifically, it performs $n$-fold (leave-one-out) cross-validation~\citep{kohavi1995} without solving the problem $n$ times by minimizing a loss function on the training data. Thus, it does not require validation data and is done highly efficiently using the low-rank projected solution~\eqref{eq:projected_sol}. 

In particular, in each iteration of the hybrid scheme, we minimize the GCV function for the projected problem~\eqref{eq:proj_eq_LS} given by
\begin{equation}\label{eq:proj_GCV}
    G_{\bfB_k, \beta \bfe_1}(\alpha) = \frac{k\| (\bfI-\bfB_k \bfB_{k,\alpha}^\dagger ) \beta \bfe_1\|_2^2}{(\text{trace}(\bfI-\bfB_k \bfB_{k,\alpha}^{\dagger}))^2}.
\end{equation}
Here, the SVD of $\bfB_k$ is performed quickly because it is small in size ($(k+1)$-by-$k$). We can then plug the SVD into~\eqref{eq:proj_GCV}. The minimization will become a simple one-dimensional problem and can be done by standard algorithms. This renders the GCV minimization effective, which needs to be done in each iteration. In principle, we can compute the GCV also for the full problem~\eqref{eq:RLS_sep}. However, this would be computationally very expensive because the full SVD of $\bfZ$ is required. For more details, see~\cite{chung2008}.
    \begin{figure}[t]
\floatconts
  {fig:hybrid_vary_para}
  {\caption{The results obtained by the hybrid method with different numbers of iterations on MNIST and CIFAR10 data. The results when the iteration number is greater than $m$ are not shown as the algorithm converges after $m$ iterations. This is because the Krylov subspace is $\mathbb{R}^{m}$, and the projected problem becomes the original problem.}}
  {\includegraphics[width=1\textwidth]{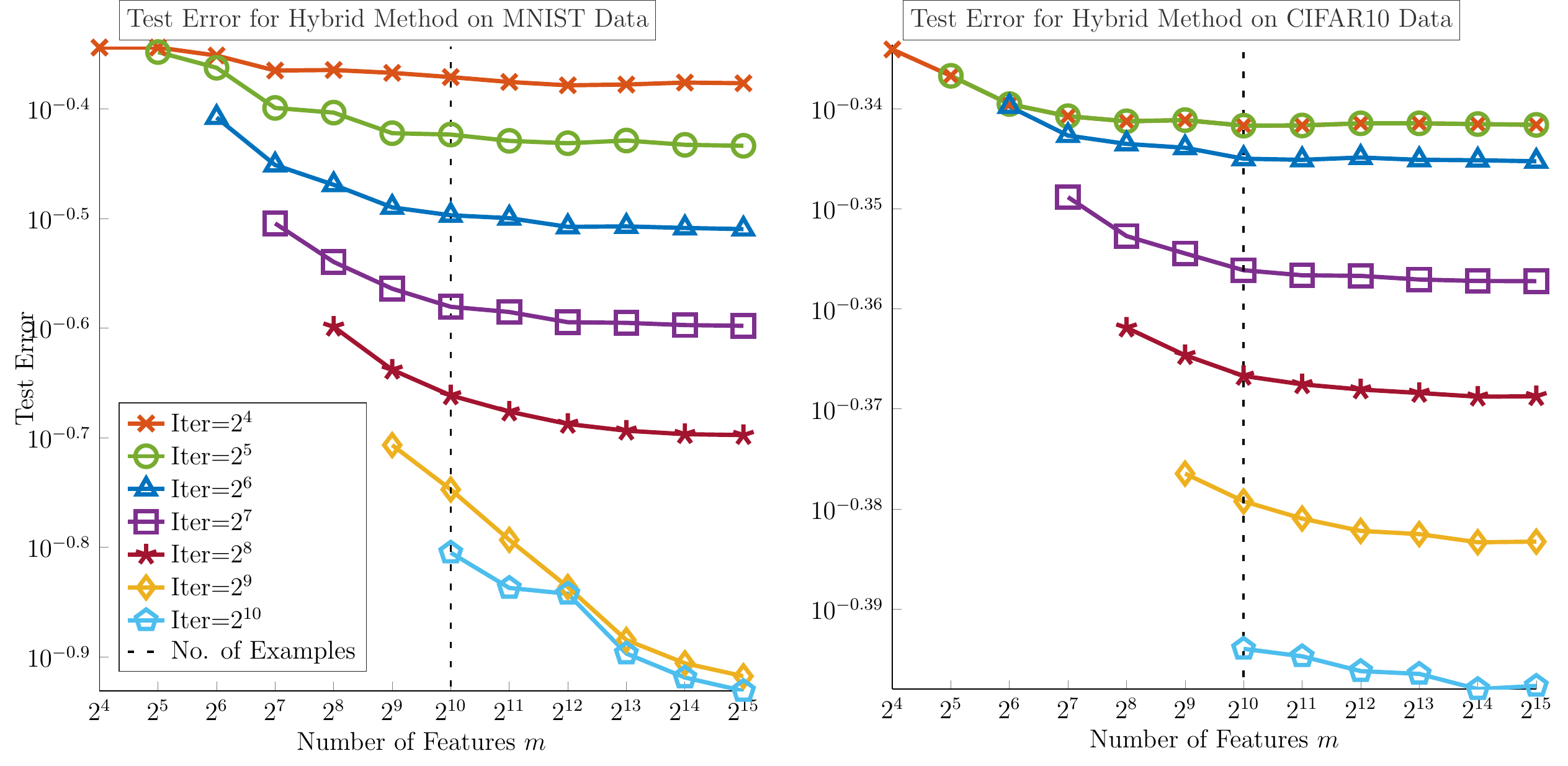}}
  \end{figure}
  
\paragraph{Advantages of the Hybrid Method} The regularization imposed by the hybrid method provides important distinct advantages over previously discussed approaches. First, in~\eqref{eq:proj_GCV}, the hybrid method performs an adaptive weight decay by dynamically selecting the hyperparameter using information from the small (but increasing) dimension Krylov subspace. Specifically, the hybrid method chooses the Tikhonov filter factors in~\eqref{eq:sol_RLS} based on the singular values of the projected problem~\eqref{eq:proj_GCV}, and these singular values are increasingly better approximations of the full dimension problem~\eqref{eq:RLS_sep}. Secondly, one can also use the GCV function value as criteria for early stopping, see~\cite{chung2008,bjorck1994}. This effectively employs a safeguard regularization. However, the Tikhonov filter factors computed in the hybrid method ensures that the computed solutions are much less sensitive to the precise stopping iteration. This combination of safeguarded regularization, automatic tuning of weight decay hyperparameters, and automatic iteration stopping criteria is very powerful.
  
\paragraph{Numerical Results} 

We apply the hybrid method to the image classification problems.
To demonstrate that the hybrid method avoids semi-convergence, we show the test error for an increasing number of iterations in \Cref{fig:hybrid_vary_para}.
For all $m$ and for both datasets, we see that increasing the number of iterations reduces the test errors until the number of iterations reaches $m$  when the bidiagonalization is exact.
In particular, the hybrid scheme avoids the double descent phenomenon when $m=n$.
 In contrast to early stopping and weight decay regularization, the hyperparameter $\alpha$ of the hybrid method is automatically chosen in each iteration.
We recommend choosing the number of iterations to match the computational budget.

\section{Comparison and Discussion}\label{sec:discussion}
In this section, we compare and discuss the numerical results achieved with the different regularization schemes.

\paragraph{Baseline} 
We optimize the weights for the early stopping and weight decay regularizers presented in Section~\ref{sec:reg} using the test data. 
It is important to emphasize that this is neither practical nor advised in realistic applications.
However, our goal is to obtain competitive baselines to compare with the hybrid scheme in which neither the test data nor some validation data is used.

We optimize the weights for each of the datasets and different widths of the RFM. 
To this end, we compute the (economic) SVD of the feature matrix $\bfZ$ and use the filter factors in~\eqref{eq:phiGF} and \eqref{eq:phiWD}, respectively, to efficiently compute the optimal weights for different choices of $t$ and $\alpha$.
Then, we minimize the test error over the hyperparameters using the one-dimensional optimization method \texttt{fminsearch} in MATLAB. 
To reduce the risk of being trapped in a suboptimal local minimum, we first evaluate the test loss at 100 points spaced equally on the logarithmic axes shown in~\Cref{fig:GF_WD} and initialize the optimization method at the hyperparameter with the lowest test loss.

\paragraph{Comparison}
We report the results for the different datasets, different $m$, and regularization schemes in~\Cref{fig:discussion_comparison}.
For the hybrid method we set the number of iterations to $\min(m,1024)$; see also~\Cref{fig:hybrid_vary_para}.
We see that the hybrid method achieves competitive test errors even though it does not use the test data set.
Remarkably, the hybrid method's solution is on par with the other schemes at the interpolation threshold.

\paragraph{Discussion}

These numerical experiments demonstrate the potential of hybrid methods to reliably train random feature models of various sizes with automatic hyperparameter tuning. 

Most practical implementations of early stopping and weight decay use parts of the training samples to tune hyperparamters using (one-fold) cross-validation.
This has the disadvantage of reducing the number of samples available during training, which generally leads to larger test error.
By contrast, the hybrid scheme implicitly performs an $n$-fold (leave-one-out) cross-validation. 

As shown in~\Cref{fig:hybrid_vary_para}, the test error of the hybrid scheme decreases with the number of iterations. 
This is in stark contrast to the gradient flow scheme (see~\Cref{fig:GF_WD}) for which semiconvergence is observed, and an adequate stopping rule is needed to avoid large generalization errors. 

In classical weight decay regularization, the learning problem has to be solved repeatedly until a weight decay parameter $\alpha$ with low validation error is found.
The hybrid scheme saves this computational overhead and automatically selects $\alpha$ in one single pass.
This is achieved by using the Lanczos bidiagionalization to evaluate the GCV function efficiently for different values of $\alpha$.

    \begin{figure}[t]
\floatconts
  {fig:discussion_comparison}
  {\caption{The results obtained by gradient flow, weight decay and the hybrid method. For gradient flow and weight decay, the optimal testing losses over time and $\alpha$, respectively, are reported. Specifically, we minimize the testing losses with respect to the hyperparameters. For the hybrid method, we determine the weight decay hyperparameters using the training data only and report the test loss with $\min(m,1024)$ iterations.}}
  {\includegraphics[width=1\textwidth]{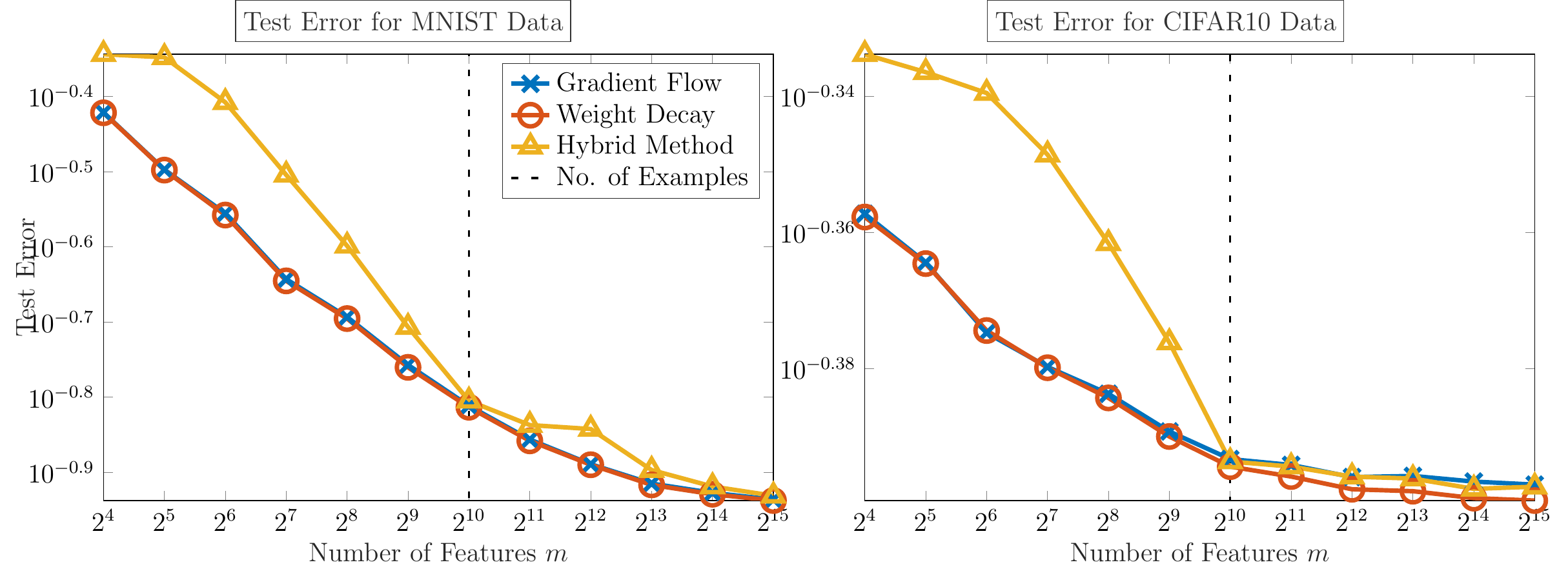}}
  \end{figure}

\section{Conclusion}\label{sec:conclusion}
We presented hybrid methods and showed in numerical experiments that they avoid the double descent phenomenon arising in the training of random feature models.
We demonstrate that the double descent phenomenon is related to the ill-posedness of the training problem.
As such, it can be overcome with early stopping and weight decay regularization; however, these techniques typically require cumbersome hyperparameter tuning, solution of multiple instances of the learning problem, and a dedicated validation set.
Hybrid methods overcome these disadvantages. 
They are computationally efficient thanks to early stopping and performing parameter search on a low-dimensional subspace. 
Further, they automatically select hyperparameters using  statistical criteria. 
While these properties have made hybrid methods increasingly popular for solving large-scale inverse problems, our paper presents the first use-case in machine learning.
In our experiments, the hybrid method performs competitively to classical regularization methods with optimally chosen weights. 
In future works, we plan to extend hybrid methods to more general learning problems, particularly with other loss functions and machine learning models. We provide our MATLAB codes for the numerical experiments at \url{https://github.com/EmoryMLIP/HybridRFM}.

\acks{KK and LR's work was supported in part by NSF award DMS 1751636 and AFOSR grant 20RT0237. JN's work was supported in part by NSF award DMS 1819042.}

\bibliography{main}






\end{document}

%% file: abbrv.tex
\newcommand{\bfB}{{\bf B}}
\newcommand{\bfC}{{\bf C}}

\newcommand{\bfI}{{\bf I}}

\newcommand{\bfK}{{\bf K}}

\newcommand{\bfP}{{\bf P}}
\newcommand{\bfQ}{{\bf Q}}

\newcommand{\bfU}{{\bf U}}
\newcommand{\bfV}{{\bf V}}
\newcommand{\bfW}{{\bf W}}

\newcommand{\bfY}{{\bf Y}}
\newcommand{\bfZ}{{\bf Z}}

\newcommand{\bfb}{{\bf b}}
\newcommand{\bfc}{{\bf c}}
\newcommand{\bfe}{{\bf e}}

\newcommand{\bfx}{{\bf x}}

\newcommand{\bfu}{{\bf u}}

\newcommand{\bfp}{{\bf p}}

\newcommand{\bff}{{\bf f}}
\newcommand{\bfv}{{\bf v}}
\newcommand{\bfw}{{\bf w}}

\DeclareMathOperator*{\argmin}{arg\,min}

%% file: preamble.tex
\usepackage{amsmath,amsfonts,amssymb}
\usepackage{xcolor}
\usepackage{tikz}
\usetikzlibrary{arrows,calc}
\usepackage{relsize}
\usepackage{pgfplots}
\usepackage{tikz}
\usetikzlibrary{plotmarks}
\usepackage{standalone}
\usepackage{graphicx}
\usepackage{cleveref}
\graphicspath{{figures/}}

